\documentclass[runningheads]{llncs}
\usepackage{makeidx}
\usepackage{graphicx}
\usepackage{grffile}
\usepackage{amsmath,amssymb} % define this before the line numbering.
\usepackage{color}

\usepackage[acronym]{glossaries}
\usepackage[caption=false]{subfig}
\usepackage{csquotes}
\usepackage{enumitem}
\usepackage{xspace}
\usepackage[table]{xcolor}
\usepackage{cleveref}

\newcommand{\etal}{\emph{et al.}\xspace}
\newcommand{\eg}{\emph{e.g.,}\xspace}
\newcommand{\ie}{\emph{i.e.,}\xspace}
\newcommand{\wrt}{\emph{w.r.t.}\xspace}

\newcommand{\xagreementfig}[3]{%
\begin{minipage}[t]{\linewidth}
\includegraphics[width=0.5\linewidth]{plots/cifar-100/class_imgs/#1.png}%
\includegraphics[width=0.5\linewidth]{plots/cifar-100/class_imgs/#2.png}\\
\centering
\tiny{#3}
\vspace{1mm}
\end{minipage}
}

\begin{document}

%%%%%%%%% ACRONYMS
\newacronym{dag}{DAG}{directed acyclic graph}
\newacronym{mlnp}{MLNP}{mandatory labeled node prediction}
\newacronym{anp}{ANP}{arbitrary node prediction}
\newacronym{cnn}{CNN}{convolutional neural network}
\newacronym{gp}{GP}{Gaussian process}

%%%%%%%%% TITLE
\pagestyle{headings}
	\mainmatter

	% Insert your submission number here
	\def\GCPR19SubNumber{30}

	% Replace with your title
	\title{Not just a matter of semantics:\\the relationship between visual and semantic similarity}

	% DO NOT MODIFY these for the draft version that is used for the
	% review process.
	\titlerunning{Not just a matter of semantics}
	\authorrunning{Clemens-Alexander Brust and Joachim Denzler}
	\author{Clemens-Alexander Brust$^1$ and Joachim Denzler$^{1,2}$}
	\institute{$^1$Computer Vision Group, Friedrich Schiller University Jena, Germany\\
	$^2$Michael Stifel Center Jena, Germany}

	\maketitle

%%%%%%%%% ABSTRACT
\begin{abstract}
Knowledge transfer, zero-shot learning and semantic image retrieval are methods that aim at improving accuracy by utilizing semantic information, \eg from WordNet. It is assumed that this information can augment or replace missing visual data in the form of
labeled training images because semantic similarity correlates with visual similarity.

This assumption may seem trivial, but is crucial for the application of such semantic methods. Any violation can cause mispredictions. Thus, it is important to examine the visual-semantic relationship for a certain target problem. In this paper, we use five different semantic and visual similarity measures each to thoroughly analyze the relationship without relying too much on any single definition.

We postulate and verify three highly consequential hypotheses on the relationship. Our results show that it indeed exists and that WordNet semantic similarity carries more information about visual similarity than just the knowledge of \enquote{different classes look different}. They suggest that classification is not the ideal application for semantic methods and that wrong semantic information is much worse than none.
\end{abstract}

%%%%%%%%% BODY TEXT

\section{Introduction}
There exist applications in which labeled training data cannot be acquired in sufficient amounts to reach the high accuracy associated
with contemporary \glspl{cnn} with millions of parameters. These include industrial \cite{Kumar2008computer,Malamas2003survey} and
medical \cite{Litjens2017survey,Salem2018recent,Thevenot2018survey} applications as well as research in other fields like wildlife
monitoring \cite{Brust2017Gorilla,Chen2014deep,Freytag2016Chimps}.
\textit{Semantic methods} such as knowledge transfer and zero-shot learning process information about
the semantic relationship between classes from databases like WordNet \cite{Miller1995WordNet}
to allow high-accuracy classification even when training data is insufficient or missing entirely \cite{Rohrbach2011Zero}.
\textit{They can only perform well when the unknown visual relationships between classes are predictable
by the semantic relationships.}

In this paper, we analyze and test this crucial assumption by
evaluating the relationship between visual and semantic similarity in a detailed and systematic manner.
%\subsection{Hypotheses}
\label{sec:exphypos}
\begin{figure}[t]
\centering
\subfloat[A deer and a forest. By taxonomy only,
their semantic similarity is weak. Visual similarity however is strong.]{
\includegraphics[width=0.22\linewidth]{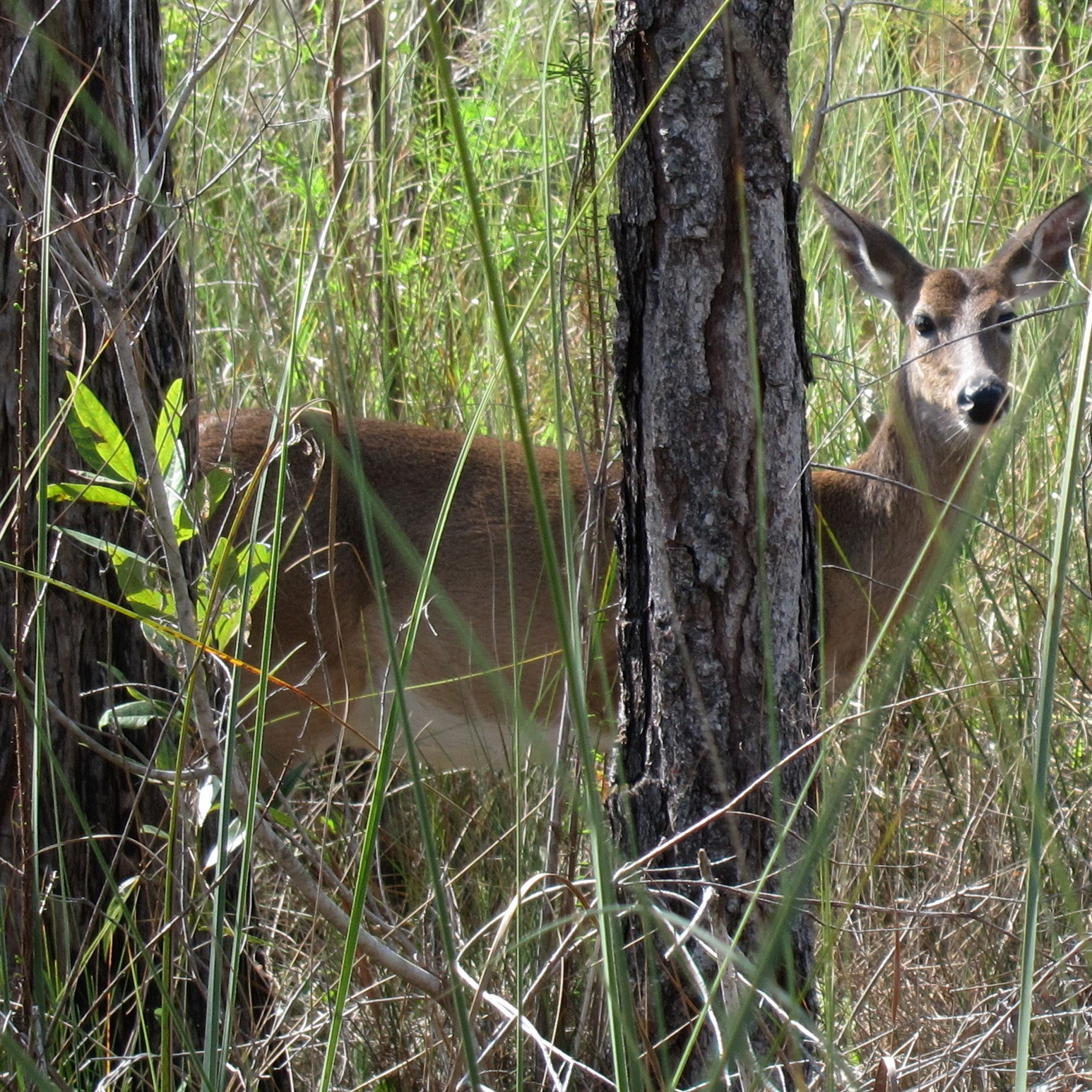}
\includegraphics[width=0.22\linewidth]{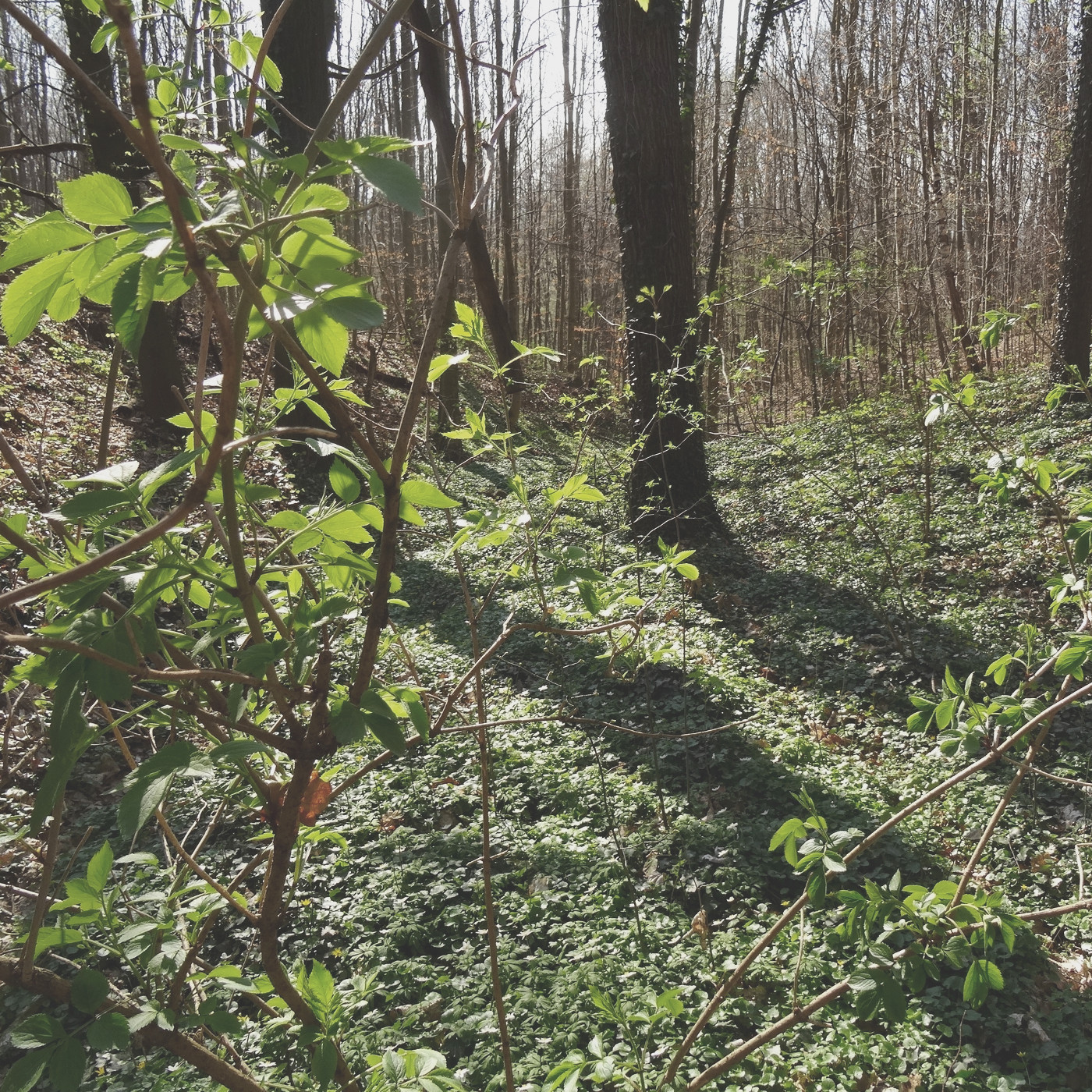}
} % \end{subfigure}\\
\hspace{0.02\linewidth}
\subfloat[An orchid and a sunflower. Their semantic similarity very strong due to them both being flowers.
The visual similarity between them is weak.]{
\includegraphics[width=0.22\linewidth]{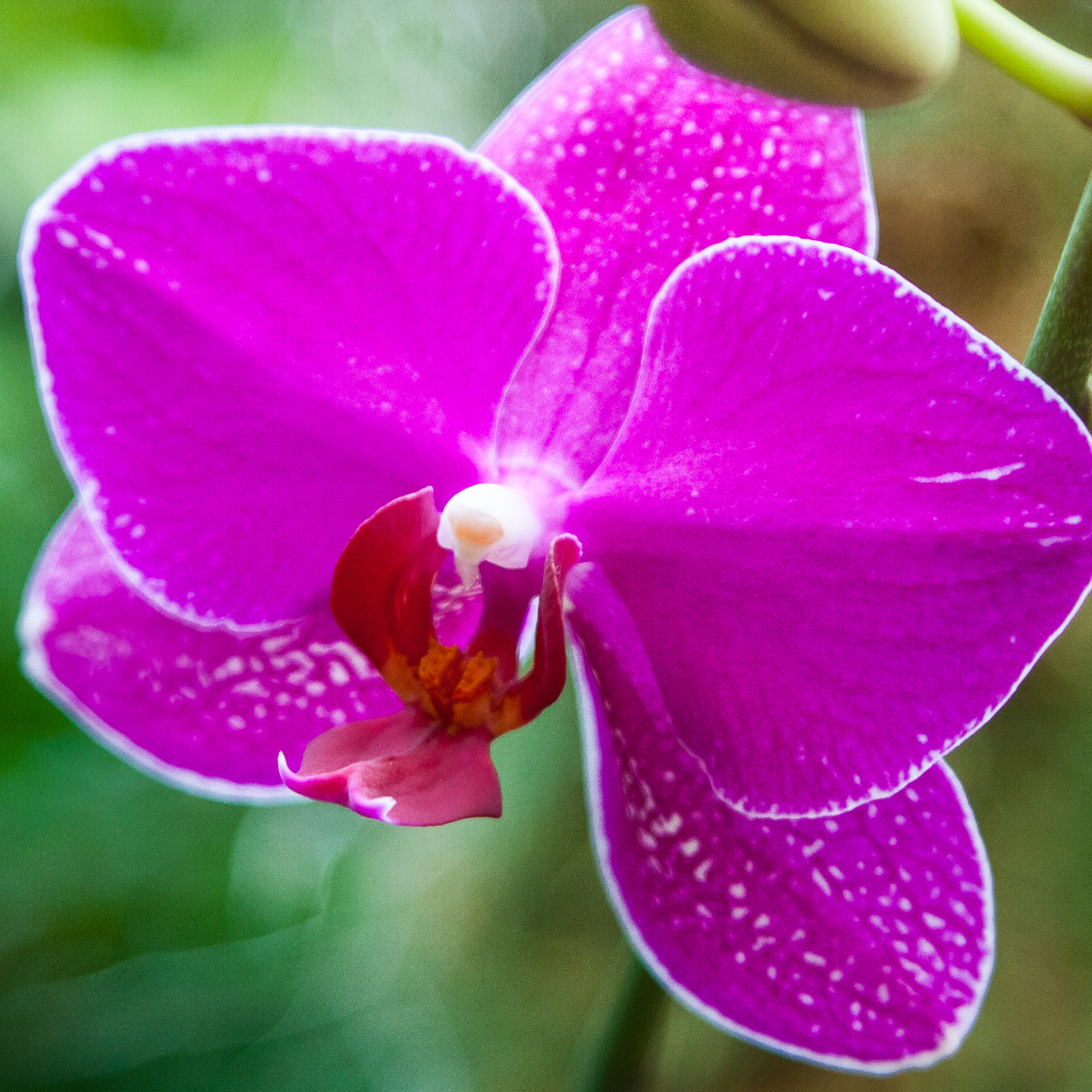}
\includegraphics[width=0.22\linewidth]{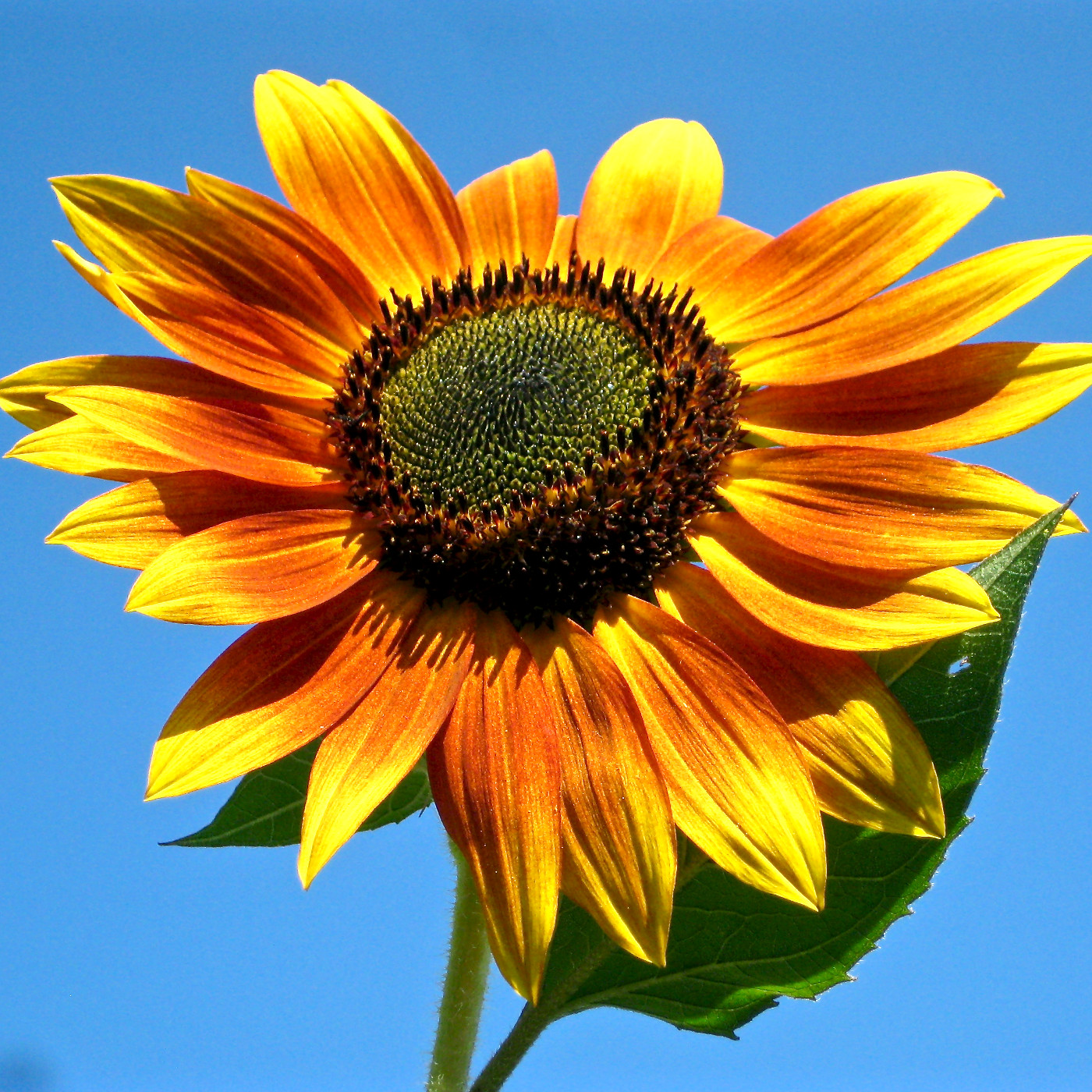}
}% \end{subfigure}
\caption{Examples of semantic-visual disagreement.}
\label{fig:teaser}
\end{figure}

To guide our analysis, we formulate three highly consequential, non-trivial hypotheses around the visual-semantic relationship.
The exact nature of the links and the similarity terms is specified in \cref{sec:expprot}.
Our first hypothesis concerns the relationship itself:

% TODO paragraphs statt enums

\paragraph{$\mathcal{H}_1$: \textbf{There is a link between visual and semantic similarity.}}
It seems trivial on the surface, but each individual component requires a proper, non-trivial definition
to ultimately make the hypothesis verifiable (see \cref{sec:expprot}).
The observed effectiveness of semantic methods suggests
that knowledge about semantic relationships is somewhat applicable in the visual domain.
However, counter-examples are easily found, \eg \cref{fig:teaser,fig:disagreement}.
Furthermore, a crude approximation of semantic similarity is already given by the
expectation that \enquote{different classes look different} (see \cref{sec:semsimprereqs}).
A similarity measure based on actual semantic knowledge should be a stronger predictor of visual
similarity than this simple baseline.

Semantic methods seek to improve accuracy and in turn reduce model confusion, but the relationship between confusion and visual similarity
is non-trivial. Insights about the low-level visual similarity may not apply to the more abstract confusion.
To cover not only largely model-free, but also also model-specific notions of visual similarity, we formulate our
second and third hypotheses:

\paragraph{$\mathcal{H}_2$: \textbf{There is a link between visual similarity and model confusion.}}
Low inter-class distance in a feature space correlates with confusion,
but it could also indicate strong visual similarity. This link depends on the selected features and
classifier. It is also affected by violations of \enquote{different classes look different} in the dataset.

\paragraph{$\mathcal{H}_3$: \textbf{There is a link between semantic similarity and model confusion.}}
This link should be investigated because it directly relates to the goal of semantic
methods, which is to reduce confusion by adding semantic information.
It \enquote{skips} the low-level visual component and as such is interesting on its own.
However, the expectation that \enquote{different classes look different} can already explain the
complete confusion matrix of a perfect classifier. We also expect it to partly explain
a real classifier's confusions. So, to consider $\mathcal{H}_3$ verified, we require semantic similarity to show
an even stronger correlation to confusion than given by this expectation.

%It is not trivial, since $\mathcal{H}_1$ and $\mathcal{H}_2$ being true
%is not a sufficient condition, \ie correlation is not neccessarily transitive.

%\paragraph{Contributions}
Our main contribution is an extensive and insightful evaluation of this relationship across five different semantic and visual
similarity measures respectively. It is based on the three aforementioned hypotheses around the relationship.
We show quantitative results measuring the agreement between individual measures and across visual and semantic similarities
as rank correlation. Moreover, we analyze special cases of agreement and disagreement qualitatively.
The results and their various implications are discussed in \cref{sec:resdisc}. They
suggest that, while the relationship exists even beyond the \enquote{different classes look different} baseline,
tasks different from classification need further research.
The semantically reductive nature of class labels suggests that semantic methods
may perform better on more complex tasks.

\subsection{Related Work}
The relationship between visual and semantic similarity has been subject of previous investigation.
In \cite{Deselaers2011Visual}, Deselaers and Ferrari consider a semantic similarity measure by Jiang and Conrath (see \cref{sec:semsiminf} and \cite{Jiang1997SemSim}) as well as category histograms, in conjunction with the ImageNet dataset.
They propose a novel distance function based on semantic as well as visual similarity
to use in a nearest neighbor setting that outperforms purely visual distance functions.
The authors also show a positive correlation between visual and semantic similarity
for their choice of similarity measures on the ImageNet dataset. Their selections
of Jiang-Conrath distance and the GIST feature descriptor are also evaluated in our work,
where we add several different methods to compare.

Bilal \etal observe the confusion matrix of a convolutional network
trained on the ImageNet-1k dataset \cite{Russakovsky2015imagenet} in \cite{bilal_convolutional_2018}.
They use visual analytics to show that characteristics of the class hierarchy can be found in the confusion matrix, a result related
to our hypothesis $\mathcal{H}_3$.

\section{Semantic Similarity}
\label{sec:semsim}
The term \textit{semantic similarity} describes the degree to which two concepts interact semantically.
A common definition requires taking into account only the taxonomical (hierarchical) relationship
between the concepts \cite[p.~10]{Harispe2015SemSim}. A more general notion is \textit{semantic relatedness},
where any type of semantic link may be considered \cite[p.~10]{Harispe2015SemSim}. Both are 
\textit{semantic measures}, which also include distances and dissimilarities \cite[p.~9]{Harispe2015SemSim}.
We adhere to these definitions in this work, specifically the hierarchical restriction of semantic similarity.

% Similarity and distances
\subsection{Prerequisites}
\label{sec:semsimprereqs}
In certain cases, it is easier to formulate a semantic measure based on hierarchical relationships as a distance
first. Such a distance $d$ between two concepts $x,y$ can be converted to a similarity by
$1/(1+d(x,y))$ \cite[p.~60]{Harispe2015SemSim}. This results in a
measure bounded by $(0,1]$, where $1$ stands for maximal similarity, \ie the distance is zero.
We will apply this rule to convert all distances to similarities in our experiments. We also apply it to
dissimilarities, which are comparable to distances, but do not fulfill the triangle inequality.

% Baseline
% TODO Lit sememb + hclass
\paragraph{Semantic Baseline}
When training a classifier without using semantic embeddings \cite{barz2019hierarchy} or hierarchical classification techniques \cite{Silla2011HClass},
there is still prior information about semantic similarity given by the classification problem itself.
Specifically, it is postulated that \enquote{classes that are different look different} (see \cref{sec:expprot}). Machine learning
can not work if this assumption is violated, \ie different classes look identical.
We encode this \enquote{knowledge} in semantic similarity measure, defined as $1$ for two
identical concepts and zero otherwise. It will serve as a baseline for comparison with all other similarities.

\subsection{Graph-based Similarities}
We can describe
a directed acyclic graph $G(C, \text{\textbf{is-a}})$ using the taxonomic relation \textbf{is-a} and the set of all concepts $C$.
The notions of semantic similarity described in this section
can be expressed using properties of $G$. The graph distance $d_G(x,y)$ between two nodes $x,y$,
which is defined as the length of the shortest path $xPy$, is an important example.
If required, we reduce the graph $G$ to a rooted tree $T$ with root $r$ by
iterating through all nodes with multiple ancestors and successively removing the edges to ancestors with the lowest
amount of successors. In a tree, we can then define the depth of a concept $x$ as $\mathfrak{d}_T(x)=d_T(r,x)$.

A simple approach is presented by
Rada \etal in \cite[p.~20]{Rada1989SemSim}, where the semantic distance between
two concepts $x$ and $y$ is defined as the graph distance $d_G(x,y)$ between one concept and the other in $G$.
% We convert this distance to a similarity using the rule described in \cref{sec:semsimprereqs}.

To make similarities comparable between different taxonomies, it may be desirable to take
the overall depth of the hierarchy into account. Resnik presents
such an approach for trees in \cite{Resnik1995SemSim}, considering the maximal depth of
$T$ and the \textit{least common ancestor} $\mathcal{L}(x,y)$.
$\mathcal{L}$ is the uniquely defined node in the shortest path between two concepts
$x$ and $y$ that is an ancestor to both \cite[p.~61]{Harispe2015SemSim}. The similarity between $x$ and $y$ is then given as \cite[p.~3]{Resnik1995SemSim}:
\begin{equation}
2 \cdot \max_{z \in C}{\mathfrak{d}_T(z)} - d_T(x,\mathcal{L}(x,y)) - d_T(y,\mathcal{L}(x,y)).
\label{eq:resnik1995}
\end{equation}

\subsection{Feature-based Similarities}
The following approaches use a set-theoretic view of semantics. The set of \textit{features}
$\phi(x)$ of a concept $x$ is usually defined as the set of ancestors $A(x)$ of $x$ \cite{Harispe2015SemSim}.
We also include $x$ itself, such that $\phi(x) = A(x) \cup \{x\}$ \cite{Sanchez2012SemSim}.

Inspired by the Jaccard coefficient, Maedche and Staab propose a
similarity measure defined as the intersection over union of the concept features of $x$ and $y$ respectively \cite[p.~4]{Maedche2001SemSim}.
This similarity is bounded by $[0,1]$, with identical concepts always resulting in $1$.

Sanchez \etal present a dissimilarity measure that represents
the ratio of distinct features to shared features of two concepts. It is defined by \cite[p.~7723]{Sanchez2012SemSim}:
\begin{equation}
\log_2 \Big(1+\frac{|\phi(x) \backslash \phi(y)| + |\phi(y) \backslash \phi(x)|}{|\phi(x) \backslash \phi(y)| + |\phi(y) \backslash \phi(x)| + |\phi(y) \cap \phi(x)|}\Big).
\label{eq:sanchez2012}
\end{equation}

\subsection{Information-based Similarities}
\label{sec:semsiminf}
Semantic similarity is also defined using the notion of informativeness of a concept, inspired by information theory.
Each concept $x$ is assigned an \textit{Information Content} (IC) $\mathcal{I}(x)$ \cite{Resnik1995SemSim,Ross1976IC}.
This can be defined using only properties of the taxonomy, \ie the graph $G$ (intrinsic IC), or
using the probability of observing the concept in corpora (extrinsic IC) \cite[p.~54]{Harispe2015SemSim}.

We use an intrinsic definition presented by Zhou \etal in \cite{Zhou2008IIC}, based on
the descendants $D(x)$:
\begin{equation}
\mathcal{I}(x) = k \cdot \Big(1- \frac{|D(x)|}{|C|} \Big) +
(1-k) \cdot \Big(\frac{\log(\mathfrak{d}_T(x))}{\log(\max_{z \in C}{\mathfrak{d}_T(z)})}\Big).
\label{eq:zhou2008}
\end{equation}
% TODO Über k schreiben

With a definition of IC, we can apply an information-based similarity
measure. Jiang and Conrath propose a semantic distance in \cite{Jiang1997SemSim}
using the notion of \textit{Most Informative Common Ancestor}
$\mathfrak{M}(x,y)$ of two concepts $x,y$. It is defined as the element in $(A(x) \cap A(y)) \cup ({x} \cap {y})$ with the highest IC \cite[p.~65]{Harispe2015SemSim}.
The distance is then defined as \cite[p.~8]{Jiang1997SemSim}:
\begin{equation}
\mathcal{I}(x) + \mathcal{I}(y) - 2 \cdot \mathcal{I}(\mathfrak{M}(x,y)).
\label{eq:jiang1997}
\end{equation}

\section{Visual Similarity}
\label{sec:vissim}
Assessing the similarity of images is not a trivial task, mostly because
the term \enquote{similarity} can be defined in many different ways.
In this section, we look at two common interpretations of visual similarity,
namely perceptual metrics and feature-based similarity measures.

\subsection{Perceptual Metrics}
Perceptual metrics are usually employed to quantify the distortion or information loss
incurred by using compression algorithms. Such methods aim to minimize the difference
between the original image and the compressed image and thereby maximize the similarity between
both. However, perceptual metrics can also be used to assess the similarity of two independent images.

%\paragraph{Mean Squared Error}
An image can be represented by an element of a high-dimensional vector space. In this case,
the Euclidean distance is a natural candidate for a dissimilarity measure. With the rule $1/(1+d)$ from \cref{sec:semsimprereqs},
the distance is transformed into a visual similarity measure.
%The distance between two images $x_1, x_2$ is zero if and only if $x_1 = x_2$. Otherwise,
%the negative distance is less than zero, fulfilling our basic requirement.
To normalize the measure \wrt image dimensions and to simplify calculations, the mean squared error (MSE)
is used. Applying the MSE to estimate image similarity has shortcomings.
For example, shifting an image by one pixel significantly changes the distances to other images,
and even its unshifted self. An alternative, but related measure
is the mean absolute difference (MAD), which we also consider in our experiments.

%\paragraph{Structural Similarity Index}
In \cite{Wang2004SSIM}, Wang \etal develop a perceptual metric called
\textit{Structural Similarity Index} to adress shortcomings of previous methods.
Specifically, they consider properties of the human visual system such that the index
better reflects human judgement of visual similarity.

We use MSE, MAD and SSIM as perceptual metrics to indicate visual similarity in our experiments.
There are better performing methods when considering human judgement, \eg \cite{Zhang2018DeepMetric}.
However, we cannot guarantee that humans always treat visuals and semantics as separate.
Therefore, we avoid further methods that are motivated by human
properties \cite{Tversky1977features,Van1996perceptual} or already incorporate
semantic knowledge \cite{Liu2007survey,Frome2013DeVise}.

\subsection{Feature-based Measures}
Features are extracted to represent images at an abstract level. Thus, distances in such
a feature space of images correspond to visual similarity in a possibly more
robust way than the aforementioned perceptual metrics. Features have inherent or learned
invariances \wrt certain transformations that should not affect the notion of visual similarity
strongly. However, learned features may also be invariant to transformations that do affect visual
similarity because they are optimized for semantic distinction. This behavior
needs to be considered when selecting abstract features to determine visual similarity.

GIST \cite{Oliva2001GIST} is an image descriptor that aims at describing a whole
scene using a small number of estimations of specific perceptual properties, such that
similar content is close in the resulting feature space. It is based on the notion
of a \textit{spatial envelope}, inspired by architecture, that can be extracted from
an image and used to calculate statistics.

For reference, we observe the confusions of five ResNet-32 \cite{He2015Res} models to
represent feature-based visual similarity on the highest level of abstraction. Because confusion is
not a symmetric function, we apply a transform $(M+M^T)/2$ to obtain a symmetric representation of the
confusion matrix.

\section{Evaluating the Relationship}
% TODO Put this back?
%\input{figures/pairwise}
%\subsection{Hypotheses}
%\subsection{Experimental Protocol}
\label{sec:expprot}
% TODO Pairwise similarities
Visual and semantic similarity are measures defined on different domains.
Semantic similarities compare concepts and visual similarities compare individual images.
To analyze a correlation, a common domain over which both can be evaluated is
essential. We propose to calculate similarities over all pairs of classes in an image
classification dataset, which can be defined for both visual and semantic similarities.
These pairwise similarities are then tested for correlation. The process is clarified
in the following:
\begin{enumerate}[wide, labelwidth=!, labelindent=0pt]

\item \textbf{Dataset.} We use the CIFAR-100 dataset \cite{Krizhevsky2009CIFAR} to verify our hypotheses.
This dataset has a scale at which all experiments take
a reasonable amount of time. Our computation times
grow quadratically with the number of classes as well as images.
Hence, we do not consider ImageNet \cite{Russakovsky2015imagenet}
or 80 million tiny images \cite{Torralba2008Tiny} despite their larger coverage
of semantic concepts.

\item \textbf{Semantic similarities.} We calculate semantic similarity measures
over all pairs of classes in the dataset. The taxonomic relation \textbf{is-a} is taken
from WordNet \cite{Miller1995WordNet} by mapping all classes in CIFAR-100 to their counterpart concepts in WordNet,
inducing the graph $G(C,\textbf{is-a})$.
Some measures are defined as distances or dissimilarities. We use the rule presented
in \cref{sec:semsimprereqs} to derive similarities.
The following measures are evaluated over all pairs of concepts $(x,y) \in C \times C$ (see \cref{sec:semsim}):
\begin{enumerate}
	\item[(S1)] Graph distance $d_G(x,y)$ as proposed by Rada \etal, see \cite[p.~20]{Rada1989SemSim}.
	\item[(S2)] Resnik's maximum depth bounded similarity, see \cref{eq:resnik1995} and \cite[p.~3]{Resnik1995SemSim}.
	\item[(S3)] Maedche and Staab similarity based on intersection over union of concept features \cite[p.~4]{Maedche2001SemSim}.
	\item[(S4)] Dissimilarity proposed by Sanchez \etal using distinct to shared features ratio,
	see \cref{eq:sanchez2012} and \cite[p.~7723]{Sanchez2012SemSim}.
	\item[(S5)] Jiang and Conrath's distance \cite[p.~8]{Jiang1997SemSim}, \cref{eq:jiang1997}, using intrinsic Information Content from \cite{Zhou2008IIC}, see \cref{eq:zhou2008}.
\end{enumerate}

\item \textbf{Visual similarities.} To estimate
a visual similarity between two classes $x$ and $y$, we calculate the similarity
of each test image of class $x$ with each test image of class $y$ and use the
average as an estimate. Again we apply the rule from \cref{sec:semsimprereqs}
for distances and dissimilarities. The process of comparing all images from one class
to all from another is performed for the following measures (see \cref{sec:vissim}):
\begin{enumerate}
	\item[(V1)] The mean squared error (MSE) between two images.
	\item[(V2)] The mean absolute difference (MAD) between two images.
	\item[(V3)] Structural Similarity Index (SSIM), see \cite{Wang2004SSIM}.
	\item[(V4)] Distance between GIST descriptors \cite{Oliva2001GIST} of images in feature space.
	\item[(V5)] Observed symmetric confusions of five ResNet-32 \cite{He2015Res} models trained on the CIFAR-100 training set.
\end{enumerate}

\item \textbf{Aggregation.} For both visual and semantic similarity,
there is more than one candidate method, \ie (S1)-(S5) and (V1)-(V5). For the following steps, we
need a single measure for each type of similarity, which we aggregate from (S1)-(S5) and (V1)-(V5) respectively.
Since each method has its merits, selecting
only one each would not be representative of the type of similarity. The output of all candidate
methods is normalized individually, such that its range is in $[0,1]$.
We then calculate the average over each type of similarity, \ie visual and semantic,
to obtain two distinct measures (S) and (V).

\item \textbf{Baselines.} A basic assumption of machine learning is that
\enquote{the domains occupied by features of different classes are separated} \cite[p.~8]{Niemann2012pattern}.
Intuitively, this should apply to the images of different classes as well.
We can then expect to predict at least some of the visual similarity between classes
just by knowing whether the classes are identical or not.
This knowledge is encoded in the \emph{semantic baseline} (SB), defined as $1$ for identical concepts and
zero otherwise (see also \cref{sec:semsimprereqs}).
We propose a second baseline, the \emph{semantic noise} (SN), where the aforementioned
pairwise semantic similarity (S) is calculated, but the concepts are permuted randomly. This
baseline serves to assess the informativeness of the taxonomic relationships.

\item \textbf{Rank Correlation} The similarity measures mentioned above are useful to
define an order of similarity, \ie whether a concept $x$ is more similar to $z$ than concept $y$.
However, it is not reasonable in all cases to interpret them in a linear fashion, especially since many are derived
from distances or dissimilarities
and all were normalized from different ranges of values and then aggregated.
We therefore test the similarities for correlation \wrt ranking,
using Spearman's rank correlation coefficient \cite{Spearman1904proof} instead
of looking for a linear relationship.

\end{enumerate}

\section{Results}
\begin{figure}[t]
\centering
\subfloat[Semantic similarities]{
\includegraphics[width=0.27\linewidth]{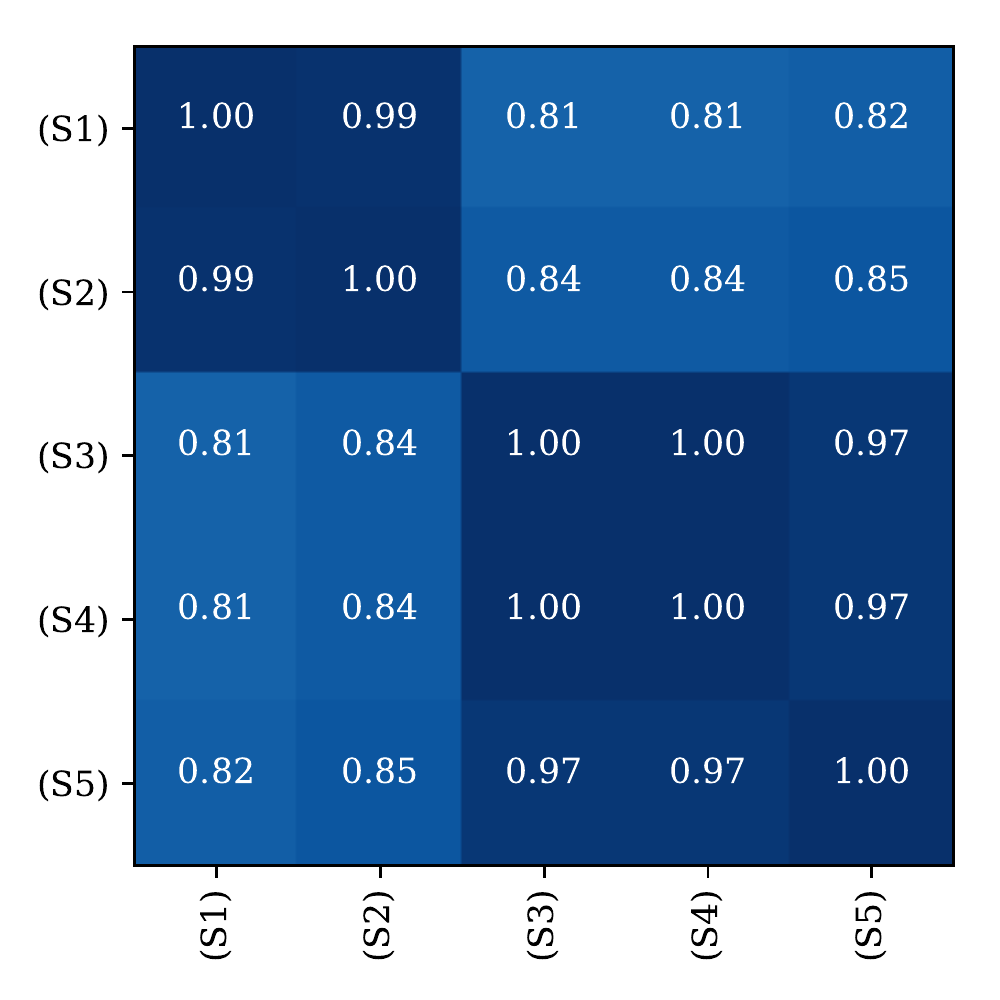}
\label{fig:rcorrsemantic}
}%
\subfloat[Visual similarities]{
\includegraphics[width=0.27\linewidth]{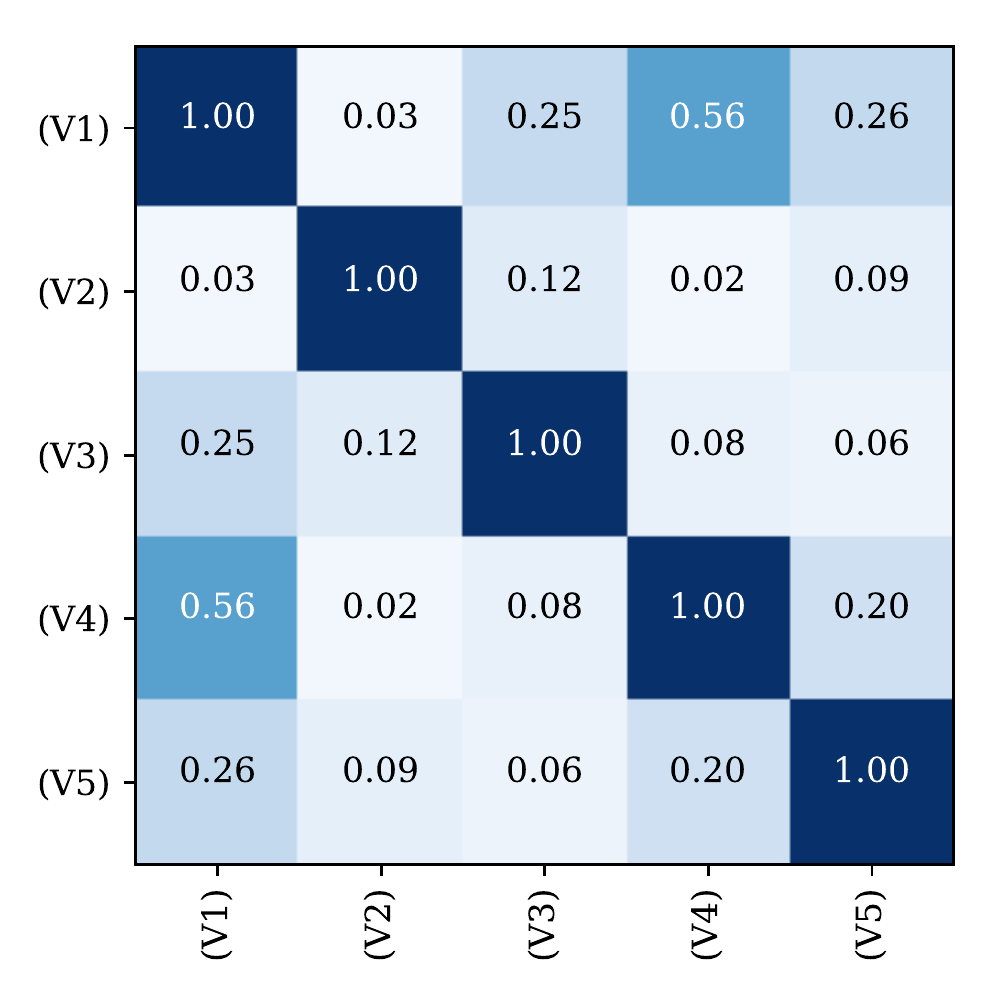}
\label{fig:rcorrvisual}
}%
\caption{Rank correlation coefficient between different similarities, grouped by semantic and
visual. $p < 0.05$ for all correlations.}
\end{figure}

In the following, we present the results of our experiments defined in the
previous section. We first examine both types of similarity individually,
comparing the five candidate methods each. Afterwards, the hypotheses
proposed in \cref{sec:exphypos} are tested. We then investigate cases
of (dis-)agreement between both types of similarity.

% TODO put this back?
% We then perform a qualitative
% analysis of extreme cases in both similarities and investigate cases
% of (dis-)agreement between them.

% Want to measure rank correlation between visual and semantic similarity
% There is not one visual similarity and one semantic similarity, there are several measures for each
% Using only one measure would not be fair, since each has its merits
% Calculate average normalized pairwise similarities between classes
\subsection{Semantic Similarities}
We first analyze the pairwise semantic similarities over all classes.
% TODO Put this back?
% \Cref{fig:pairwisesem} shows the average semantic similarity (S) as specified in
% \cref{sec:expprot}. The classes on the axes are ordered by a depth first search
% through the class hierarchy, yielding clearly visible artifacts of the graph structure.
Although we consider semantic similarity to be a single measure when verifying our
hypotheses, studying the correlation between our candidate methods (S1)-(S5) is
also important. While of course affected by our selection,
it reflects upon the degree of agreement between several experts in the domain.
\Cref{fig:rcorrsemantic} visualizes the correlations. The graph-based
methods (S1) and (S2) agree more strongly with each other than with the rest. The same
is true of feature-based methods (S3) and (S4), which show the strongest correlation.
The \emph{inter-agreement} $R$, calculated by taking the average of all correlations except for
the main diagonal, is \textbf{0.89}. This is a strong agreement and suggests that the order of
similarity between concepts can be, for the most part, considered representative of a universally
agreed upon definition (if one existed). At the same time, one needs to consider that
all methods utilize the same WordNet hierarchy.

\paragraph{Baselines}
Our semantic baseline (SB, see \cref{sec:expprot}) encodes the basic knowledge
that different classes look different. This property should also be fulfilled by the average semantic
similarity (S, see \cref{sec:expprot}). We thus expect there to be at least some correlation.
The rank correlation between our average semantic similarity (S) and the semantic baseline (SB)
is $0.17$ with $p < 0.05$. This is a weak correlation compared to the strong inter-agreement of 0.89,
which suggests that the similarities (S1)-(S5) are vastly more complex than (SB), but at the same time
have a lot in common. As a second baseline we test the semantic noise (SN, see \cref{sec:expprot}).
It is not correlated with (S) at $\rho=0.01, p > 0.05$, meaning that the taxonomic relationship
strongly affects (S). If it did not, the labels could be permuted without changing the pairwise similarities.

\subsection{Visual Similarities}
% TODO Appendix?
% The average visual similarity (V) as estimated over all classes is shown
% in \cref{fig:pairwisevis}. For reference, we show the symmetric confusion matrix
% (see \cref{sec:expprot}) in \cref{fig:pairwiseconf}. Comparing (V) to (S), the
% graph structure is less visible. In the confusion matrix however, the artifacts
% are more present.

Intuitively, visual similarity
is a concept that is hard to define clearly and uniquely.
Because we selected very different approaches with very different ideas and motivations behind them,
we expect the agreement between (V1)-(V5) to be weak. \Cref{fig:rcorrvisual} shows the rank
correlations between each candidate method. The agreement is strongest between the mean squared error (V1)
and the GIST feature distance (V4). Both are L2 distances, but calculated in separate domains,
highlighting the strong nonlinearity and complexity of image descriptors. The
inter-agreement is very weak at $R=\textbf{0.17}$. The results confirms our intuitions that visual
similarity is very hard to define in mathematical terms. There is also no body of knowledge that all methods
use in the visual domain like WordNet provides for semantics.

\subsection{Hypotheses}
\label{sec:reshypo}
\begin{figure}[t]
\centering
\includegraphics[width=0.33\linewidth]{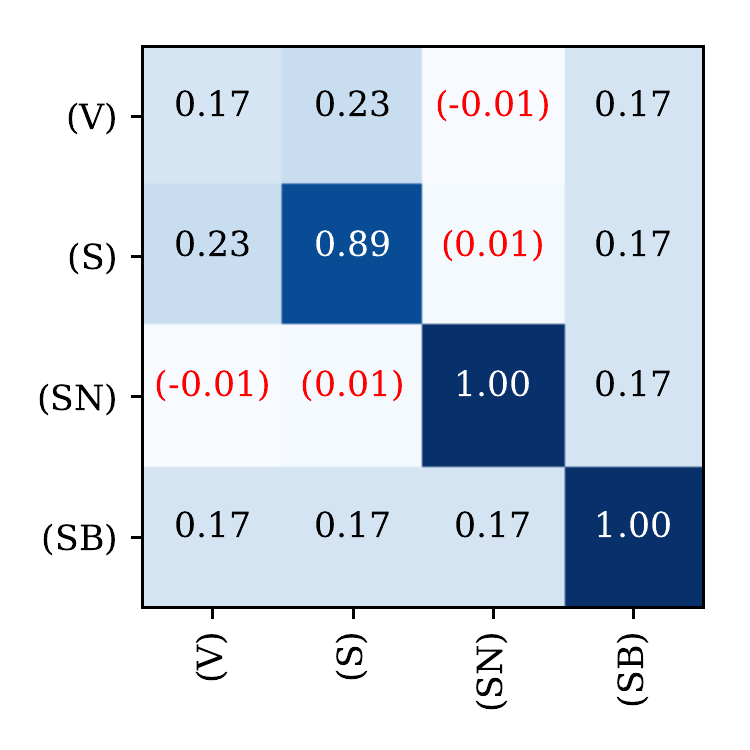}
\caption{Rank correlation coefficient between different types of similarities, grouped by semantic and
visual. $p < 0.05$ except for numbers in parentheses. Main diagonal represents inter-agreement $R$.
Similarities: (V) -- visual, (S) -- semantic, (SN) -- semantic noise, (SB) -- semantic baseline.}
\label{fig:theone}
\end{figure}

To give a brief overview, the rank correlations between the different components of $\mathcal{H}_1$-$\mathcal{H}_3$ are
shown in \cref{fig:theone}. In the following, we give our results \wrt the individual hypotheses. They are
discussed further in \cref{sec:resdisc}.

\paragraph{$\mathcal{H}_1$: There is a link between visual and semantic similarity.}
Using the definitions from \cref{sec:expprot} including the semantic baseline (SB), we can examine the
respective correlations. The rank correlation between (V) and (S) is \textbf{0.23}, $p < 0.05$,
indicating a link. Before we consider the hypothesis verified,
we also evaluate what fraction of (V) is already explained by the semantic baseline (SB) as per our
condition given in \cref{sec:expprot}.
The rank correlation between (V) and (SB) is 0.17, $p < 0.05$, which is a weaker link than between (V) and (S).
Additionally, (V) and (SN) are not correlated, illustrating that the wrong semantic knowledge can be worse
than none. 
Thus, we can \textbf{verify} $\mathcal{H}_1$.

\paragraph{$\mathcal{H}_2$: There is a link between visual similarity and model confusion.}
Since model confusion as (V5) is a contributor to average visual similarity (V), we consider only
(V-), comprised of (V1)-(V4) for this hypothesis. The rank correlation between (V-) and
the symmetric model confusion (V5) %, see \cref{fig:pairwiseconf})
is \textbf{0.21}, $p < 0.05$.
Consequently, $\mathcal{H}_2$ is also \textbf{verified}.

\paragraph{$\mathcal{H}_3$: There is a link between semantic similarity and model confusion.}
Here we evaluate the relationship between (S) and the symmetric confusion matrix (V5) %, see \cref{fig:pairwiseconf})
as defined
in \cref{sec:expprot}.
(S) should offer more information about where confusions occur than the baseline (SB) to consider $\mathcal{H}_3$ verified.
The rank correlation between (V5) and (S) is \textbf{0.39}, $p < 0.05$,
while (V5) and (SB) are only correlated at $\rho = 0.21$, $p < 0.05$,
 meaning that
$\mathcal{H}_3$ is \textbf{verified}, too.

See \cref{sec:resdisc} for a discussion of possible consequences.
% - Always a good thing? Probably

% TODO Was "Special cases"
\subsection{Agreement and Disagreement}
\begin{figure}[t]
\centering
\subfloat[Most agreed upon]{
\begin{minipage}[t]{0.46\linewidth}
  \centering
  \xagreementfig{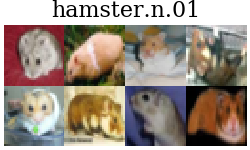}{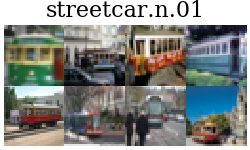}{Rank diff.: 0.01\%, sem./vis. sim.: 0.13 / 0.18}\\
  \xagreementfig{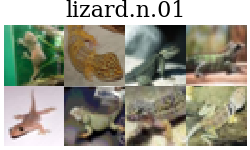}{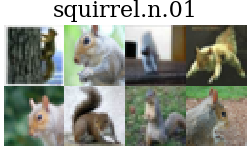}{Rank diff.: 0.01\%, sem./vis. sim.: 0.34 / 0.23}\\
  \xagreementfig{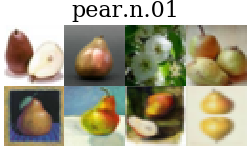}{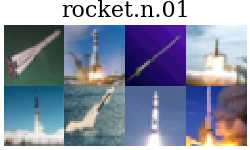}{Rank diff.: 0.01\%, sem./vis. sim.: 0.15 / 0.19}
  \label{fig:qspecagree}
\end{minipage}
}%
\hspace{0.02\linewidth}
\subfloat[Least agreed upon]{
\begin{minipage}[t]{0.46\linewidth}
  \centering
  \xagreementfig{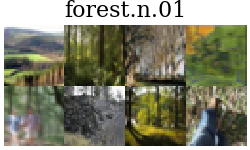}{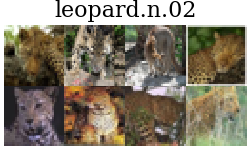}{Rank diff.: 97.23\%, sem./vis. sim.: 0.04 / 0.25}\\
  \xagreementfig{forest.n.01}{squirrel.n.01}{Rank diff.: 96.01\%, sem./vis. sim.: 0.06 / 0.25}\\
  \xagreementfig{forest.n.01}{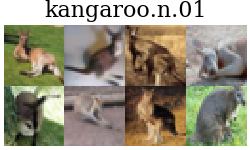}{Rank diff.: 95.53\%, sem./vis. sim.: 0.06 / 0.25}
  \label{fig:qspecdisagree}
\end{minipage}
}%
\caption{CIFAR-100 classes selected by highest
and lowest ranking agreement between visual and semantic similarity
measures as defined in \cref{sec:expprot}.}

\label{fig:disagreement}
\end{figure}

To further analyze the the correlation, we examine specific cases of very strong agreement or disagreement.
\Cref{fig:disagreement} shows these extreme cases. We determine agreement based on ranking, so the
most strongly agreed upon pairs (see \cref{fig:qspecagree}) still show different absolute similarity numbers. Interestingly,
they are not cases of extreme similarities. It suggests that even weak disagreements are more likely to be found at similarities close to the boundaries.
When investigating strong disagreement as shown in \cref{fig:qspecdisagree},
there are naturally extreme values to be found. All three pairs involve \texttt{forest.n.01}, which was
also a part of the second least semantically similar pair. Its partners are all animals which usually have
a background visually similar to a forest, hence the strong disagreement.
However, the low semantic similarity is possibly an artifact of reducing a whole image to
a single concept.

\subsection{Discussion}
\label{sec:resdisc}
\paragraph{$\mathcal{H}_1$: There is a link between visual and semantic similarity.}
The relationship is stronger than a simple baseline,
but weak overall at $\rho=0.23$ vs $\rho=0.17$. This should be considered when employing methods where
visuals and semantics interact, \eg in knowledge transfer. Failure cases such as in \cref{fig:qspecdisagree}
can only be found when labels are known, which has implications for real-life applications of semantic methods.
As labels are unknown or lack visual examples, such cases are not predictable beforehand.
This poses problems for applications that rely on accurate classification such as safety-critical
equipment or even research in other fields consuming model predictions.
A real-world example is wildlife conservationists relying on statistics
from automatic camera trap image classification to draw conclusions
on biodiversity. That semantic similarity of randomly permuted classes is not correlated with visual similarity
at all, while the baseline is, suggests that wrong semantic knowledge can be much the worse than no knowledge.

\paragraph{$\mathcal{H}_2$: There is a link between visual similarity and model confusion.}
Visual similarity is defined on a low level for $\mathcal{H}_2$. As such, it should not cause model confusion
by itself. On the one hand, the model can fail to generalize and cause an avoidable confusion.
On the other hand, there may be an issue with the dataset. The test set may be sampled from a
different distribution than the training set. It may also violate the postulate that different classes
look different by containing the same or similar images across classes.

\paragraph{$\mathcal{H}_3$: There is a link between semantic similarity and model confusion.}
Similar to $\mathcal{H}_1$, it suggests that semantic methods could be applied to our data,
but maybe not in general because failure cases are unpredictable. However, it implies a stronger
effectiveness than $\mathcal{H}_1$ at $\rho=0.39$ vs. the baseline at $\rho=0.21$.
We attribute this to the model's capability of abstraction. It aligns with the idea of
taxonomy, which is based on repeated abstraction of concepts. Using a formulation that optimizes semantic
similarity instead of cross-entropy (which would correspond to the semantic baseline) could help
in our situation. It may still not generalize to other settings and
any real-world application of such methods should be verified with at least a small test set.

\paragraph{Qualitative} Some failures or disagreements may not be a result of the relationship itself, but of its application to image classification. The example from \cref{fig:teaser} is valid when
the whole image is reduced to a single concept. Still, the agreement between visual and
semantic similarity may increase when the image is described in a more holistic fashion.
While \enquote{deer} and \enquote{forest} as nouns are taxonomically only loosely related,
the descriptions \enquote{A deer standing in a forest, partially occluded by a tree and tall grass} and
\enquote{A forest composed of many trees and bushes, with the daytime sky visible} already
appear more similar, while those descriptions are still missing some of the images' contents.
This suggests that more complex tasks stand to benefit more
from semantic methods.

In further research, not only nouns should be considered, but 
also adjectives, decompositions of objects into parts as well as
working with a more general notion of semantic relatedness
instead of simply semantic similarity. Datasets like Visual Genome \cite{Krishna2017VG}
offer more complex annotations mapped to WordNet concepts that could be subjected
to further study.
% However, tasks much more complex than hierarchical image classification
% on a semantic level lack a compelling real-world application to the best of our knowledge.
%
\section{Conclusion}
We present results of a comprehensive evaluation of semantic similarity measures and their
correlation with visual similarities. We measure against the simple prior knowledge of different classes having different visuals.
Then, we show that the relationship between semantic similarity, as calculated from WordNet \cite{Miller1995WordNet} using
five different methods, and visual similarity, also represented by five measures, is more meaningful than that. %, \ie can not be explained by just the baseline.
Furthermore, inter-agreement measures suggest that
semantic similarity has a more agreed upon definition than visual similarity, although
both concepts are based on human perception.

The results indicate that further research, especially into tasks different from image classification
is warranted because of the semantically reductive nature of image labels.
It may restrict the performance of semantic methods unneccessarily.
It is
possible that the relationship between semantic and visual similarity is much stronger
when the semantics of an image are better approximated.

{\small
\bibliographystyle{splncs03}
\bibliography{paper}

\begin{thebibliography}{10}
\providecommand{\url}[1]{\texttt{#1}}
\providecommand{\urlprefix}{URL }

\bibitem{barz2019hierarchy}
Barz, B., Denzler, J.: Hierarchy-based image embeddings for semantic image
  retrieval. In: 2019 IEEE Winter Conference on Applications of Computer Vision
  (WACV). pp. 638--647. IEEE (2019)

\bibitem{bilal_convolutional_2018}
Bilal, A., Jourabloo, A., Ye, M., Liu, X., Ren, L.: Do convolutional neural
  networks learn class hierarchy?  24(1),  152--162

\bibitem{Van1996perceptual}
Van~den Branden~Lambrecht, C.J., Verscheure, O.: Perceptual quality measure
  using a spatiotemporal model of the human visual system. In: Digital Video
  Compression: Algorithms and Technologies 1996. vol. 2668, pp. 450--462.
  International Society for Optics and Photonics (1996)

\bibitem{Brust2017Gorilla}
Brust, C.A., Burghardt, T., Groenenberg, M., Käding, C., Kühl, H., Manguette,
  M., Denzler, J.: Towards automated visual monitoring of individual gorillas
  in the wild. In: International Conference on Computer Vision Workshop
  (ICCV-WS) (2017)

\bibitem{Chen2014deep}
Chen, G., Han, T.X., He, Z., Kays, R., Forrester, T.: Deep convolutional neural
  network based species recognition for wild animal monitoring. In: Image
  Processing (ICIP), 2014 IEEE International Conference on. pp. 858--862. IEEE
  (2014)

\bibitem{Deselaers2011Visual}
Deselaers, T., Ferrari, V.: Visual and semantic similarity in imagenet. In:
  Computer Vision and Pattern Recognition (CVPR), 2011 IEEE Conference on. pp.
  1777--1784. IEEE (2011)

\bibitem{Freytag2016Chimps}
Freytag, A., Rodner, E., Simon, M., Loos, A., Kühl, H., Denzler, J.:
  Chimpanzee faces in the wild: Log-euclidean cnns for predicting identities
  and attributes of primates. In: German Conference on Pattern Recognition
  (GCPR). pp. 51--63 (2016)

\bibitem{Frome2013DeVise}
Frome, A., Corrado, G.S., Shlens, J., Bengio, S., Dean, J., Ranzato, M.A.,
  Mikolov, T.: Devise: A deep visual-semantic embedding model. In: Burges,
  C.J.C., Bottou, L., Welling, M., Ghahramani, Z., Weinberger, K.Q. (eds.)
  Advances in Neural Information Processing Systems 26, pp. 2121--2129. Curran
  Associates, Inc. (2013)

\bibitem{Harispe2015SemSim}
Harispe, S., Ranwez, S., Janaqi, S., Montmain, J.: Semantic similarity from
  natural language and ontology analysis. Synthesis Lectures on Human Language
  Technologies  8(1),  1--254 (2015)

\bibitem{He2015Res}
He, K., Zhang, X., Ren, S., Sun, J.: Deep residual learning for image
  recognition. In: Computer Vision and Pattern Recognition (CVPR) (2016)

\bibitem{Jiang1997SemSim}
Jiang, J.J., Conrath, D.W.: Semantic similarity based on corpus statistics and
  lexical taxonomy. In: arXiv preprint arXiv:cmp-lg/9709008 (1997)

\bibitem{Krishna2017VG}
Krishna, R., Zhu, Y., Groth, O., Johnson, J., Hata, K., Kravitz, J., Chen, S.,
  Kalantidis, Y., Li, L.J., Shamma, D.A., et~al.: Visual genome: Connecting
  language and vision using crowdsourced dense image annotations. International
  Journal of Computer Vision (IJCV)  123(1),  32--73 (2017)

\bibitem{Krizhevsky2009CIFAR}
Krizhevsky, A., Hinton, G.: Learning multiple layers of features from tiny
  images. Tech. Rep.~4, University of Toronto (2009)

\bibitem{Kumar2008computer}
Kumar, A.: Computer-vision-based fabric defect detection: A survey. IEEE
  transactions on industrial electronics  55(1),  348--363 (2008)

\bibitem{Litjens2017survey}
Litjens, G., Kooi, T., Bejnordi, B.E., Setio, A.A.A., Ciompi, F., Ghafoorian,
  M., Van Der~Laak, J.A., Van~Ginneken, B., S{\'a}nchez, C.I.: A survey on deep
  learning in medical image analysis. Medical image analysis  42,  60--88
  (2017)

\bibitem{Liu2007survey}
Liu, Y., Zhang, D., Lu, G., Ma, W.Y.: A survey of content-based image retrieval
  with high-level semantics. Pattern recognition  40(1),  262--282 (2007)

\bibitem{Maedche2001SemSim}
Maedche, A., Staab, S.: Comparing ontologies-similarity measures and a
  comparison study. Tech. rep., Institute AIFB, University of Karlsruhe (2001)

\bibitem{Malamas2003survey}
Malamas, E.N., Petrakis, E.G., Zervakis, M., Petit, L., Legat, J.D.: A survey
  on industrial vision systems, applications and tools. Image and vision
  computing  21(2),  171--188 (2003)

\bibitem{Miller1995WordNet}
Miller, G.A.: Wordnet: A lexical database for english. Commun. ACM  38(11),
  39--41 (Nov 1995)

\bibitem{Niemann2012pattern}
Niemann, H.: Pattern Analysis. Springer Series in Information Sciences,
  Springer Berlin Heidelberg (2012),
  \url{https://books.google.de/books?id=mdOoCAAAQBAJ}

\bibitem{Oliva2001GIST}
Oliva, A., Torralba, A.: Modeling the shape of the scene: A holistic
  representation of the spatial envelope. International journal of computer
  vision  42(3),  145--175 (2001)

\bibitem{Rada1989SemSim}
Rada, R., Mili, H., Bicknell, E., Blettner, M.: Development and application of
  a metric on semantic nets. IEEE transactions on systems, man, and cybernetics
   19(1),  17--30 (1989)

\bibitem{Resnik1995SemSim}
Resnik, P.: Using information content to evaluate semantic similarity in a
  taxonomy. In: arXiv preprint arVix:cmp-lg/9511007 (1995)

\bibitem{Rohrbach2011Zero}
Rohrbach, M., Stark, M., Schiele, B.: Evaluating knowledge transfer and
  zero-shot learning in a large-scale setting. In: Computer Vision and Pattern
  Recognition (CVPR), 2011 IEEE Conference on. pp. 1641--1648. IEEE (2011)

\bibitem{Ross1976IC}
Ross, S.M.: A first course in probability. Macmillan (1976)

\bibitem{Russakovsky2015imagenet}
Russakovsky, O., Deng, J., Su, H., Krause, J., Satheesh, S., Ma, S., Huang, Z.,
  Karpathy, A., Khosla, A., Bernstein, M., et~al.: Imagenet large scale visual
  recognition challenge. International Journal of Computer Vision (IJCV)
  115(3),  211--252 (2015)

\bibitem{Salem2018recent}
Salem, M.A.M., Atef, A., Salah, A., Shams, M.: Recent survey on medical image
  segmentation. In: Computer Vision: Concepts, Methodologies, Tools, and
  Applications, pp. 129--169. IGI Global (2018)

\bibitem{Sanchez2012SemSim}
S{\'a}nchez, D., Batet, M., Isern, D., Valls, A.: Ontology-based semantic
  similarity: A new feature-based approach. Expert systems with applications
  39(9),  7718--7728 (2012)

\bibitem{Silla2011HClass}
Silla, C.N., Freitas, A.A.: A survey of hierarchical classification across
  different application domains. Data Mining and Knowledge Discovery  22(1),
  31--72 (Jan 2011)

\bibitem{Spearman1904proof}
Spearman, C.: The proof and measurement of association between two things. The
  American journal of psychology  15(1),  72--101 (1904)

\bibitem{Thevenot2018survey}
Thevenot, J., L{\'o}pez, M.B., Hadid, A.: A survey on computer vision for
  assistive medical diagnosis from faces. IEEE journal of biomedical and health
  informatics  22(5),  1497--1511 (2018)

\bibitem{Torralba2008Tiny}
Torralba, A., Fergus, R., Freeman, W.T.: 80 million tiny images: A large data
  set for nonparametric object and scene recognition. Transactions on Pattern
  Analysis and Machine Intelligence (PAMI)  30(11),  1958--1970 (2008)

\bibitem{Tversky1977features}
Tversky, A.: Features of similarity. Psychological review  84(4),  327 (1977)

\bibitem{Wang2004SSIM}
Wang, Z., Bovik, A.C., Sheikh, H.R., Simoncelli, E.P.: Image quality
  assessment: from error visibility to structural similarity. IEEE Transactions
  on Image Processing  13(4),  600--612 (April 2004)

\bibitem{Zhang2018DeepMetric}
Zhang, R., Isola, P., Efros, A.A., Shechtman, E., Wang, O.: The unreasonable
  effectiveness of deep features as a perceptual metric. In: arXiv preprint
  arXiv:1801.03924

\bibitem{Zhou2008IIC}
Zhou, Z., Wang, Y., Gu, J.: A new model of information content for semantic
  similarity in wordnet. In: Future Generation Communication and Networking
  Symposia, 2008. FGCNS'08. Second International Conference on. vol.~3, pp.
  85--89. IEEE (2008)

\end{thebibliography}
}

\end{document}